\documentclass{article}

\PassOptionsToPackage{numbers, compress}{natbib}

\usepackage[dblblindworkshop, final]{neurips_2026}
\workshoptitle{Human-AI Coevolution: Measuring Human-Agent Teams in the Agentic Era}

\usepackage{amsmath}
\usepackage{graphicx}
\usepackage[utf8]{inputenc} 
\usepackage[T1]{fontenc}   
\usepackage{hyperref}
\hypersetup{
    hidelinks
}                        
\usepackage{url}            
\usepackage{booktabs}       
\usepackage{amsfonts}       
\usepackage{nicefrac}       
\usepackage{microtype}      
\usepackage{xcolor}         
\usepackage{float}
\usepackage{multirow}

\title{Cognitively Constrained Partners for Zero-Shot Coordination in Real-Time Partitioned Control}

\author{%
  Ash Simmons \\
  University of Exeter, UK \\
  \texttt{as1798@exeter.ac.uk} \\
  \And
  George De Ath \\
  University of Exeter, UK \\
  \texttt{g.de.ath@exeter.ac.uk} \\
  \And
  Richard M.~Everson \\
  University of Exeter, UK \\
  \texttt{r.m.everson@exeter.ac.uk} \\
}

\begin{document}

\maketitle

\begin{abstract}
  
Effective human-machine teaming requires agents that coordinate with unpredictable human partners. Reinforcement learning from live human feedback or large behavioural datasets is expensive, while purely simulated training overfits to artificial conventions and fails at Zero-Shot Coordination with real people. We propose a framework for training teaming agents against Cognitively Constrained Partners: simulated partners whose diversity emerges from procedural constraints grounded in human physiological literature, specifically reaction delays and perceptual limitations, rather than from reward shaping or historical training checkpoints. In a modified Lunar Lander environment, we report preliminary evidence that these partners play more like real people than unconstrained agents do, and that agents trained against them match the strongest cross-play baselines while being the only ones whose performance does not depend on how their training partners were generated. We also describe a human-participant study designed so that its rankings can be compared directly against cross-play, testing how well that standard evaluation predicts real human-agent teaming.
\end{abstract}

\section{Introduction and Motivation}
\label{sec:intro}

Effective human-machine teaming (HMT) requires autonomous agents which not only effectively maximise a reward function in isolation, but maintain optimality when coordinating with unpredictable human partners. Traditionally, training Reinforcement Learning (RL) agents for cooperation relies on human-in-the-loop feedback or extensive behavioural datasets; these approaches are often prohibitively expensive and create operational bottlenecks.

To bypass this, recent approaches pivot to purely simulated training, such as multi-agent or automated population play. These generate diverse policies efficiently and without human data, but, lacking human grounding, consistently overfit to arbitrary artificial conventions. Consequently, when these agents are deployed alongside real human partners under the conditions of Zero-Shot Coordination (ZSC) -- collaboration with a novel partner without prior interaction -- their brittle policies tend to collapse. Contemporary methods bridge this gap either by cloning human datasets, shifting the cost of collection to pre-training, or by drawing diversity from self-play checkpoints~\citep{strouse2021collaborating}, whose logic still differs fundamentally from a human teammate's.

In continuous settings, human-machine teaming is usually posed as shared control, with the human and the agent holding overlapping authority over the same task space~\citep{schaff2020residual, yan2024humancentered, reddy2018shared}. Blended assistance and corrective nudges reduce error effectively, but cast the agent as a subordinate safety net rather than a partner. Dividing the task instead runs into boundaries that are traded dynamically: either unwilling to bear the cognitive overhead of predicting the agent or lacking trust in its ability, human operators reclaim large portions of the task and circumvent the teaming process entirely~\citep{derooij2024flight, jonesjang2022how}, reducing a multi-agent system to a single-actor workflow.

Addressing these challenges, this paper presents an RL architecture deployed within a time-continuous, strictly partitioned control environment, a domain which remains under-explored as an RL benchmark for ZSC. By enforcing a rigid division of control wherein the human and the agent operate over mutually exclusive degrees of freedom, the framework eliminates the possibility of human domination while preventing the agent from being reduced to a purely assistive tool. This forced interdependence requires both partners to continuously coordinate their distinct inputs, facilitating genuine co-execution analogous to that required in real-world systems such as search and rescue, air traffic control, and robotic surgical assistance. 

To achieve this without the vast datasets required in behavioural cloning (BC) or the brittle assumptions of Fictitious Co-Play (FCP), we propose a framework centred on partners limited by the same cognitive constraints as people. Teaming performance is evaluated against Multi-Agent Reinforcement Learning (MARL), Population Based Training (PBT)~\citep{jaderberg2017population}, and FCP~\citep{strouse2021collaborating} baselines, both under cross-play with held-out partners and, in a forthcoming study, alongside human players. This pairing tests whether cross-play, the standard measure of ZSC, actually predicts coordination with real people. \looseness = -1

\section{Related Work}

While continuous partitioned control has been modelled extensively in physical systems, traditional solutions rely on computationally rigid frameworks.~\citet{javdani2018shared} managed continuous interdependence in a high-degree-of-freedom robotic arm by formulating shared autonomy as a Partially Observable Markov Decision Process, but this requires exhaustive prior knowledge of the transition dynamics and a predefined set of goals, restricting scalability where ad-hoc coordination is required.~\citet{reddy2018shared} instead demonstrated model-free Deep RL for continuous shared autonomy in a Lunar Lander environment, assisting operators without prior knowledge of the dynamics or the user's intent. That work focused on blended control, where the agent may override or modify the human's inputs to ensure safety; our architecture instead enforces strict partitioned control, preventing the agent from overriding suboptimal human play.

Self-play and population play agents coordinate well with one another, but only because each is accustomed to partners performing at a similar level of optimality to itself.~\citet{carroll2019utility} showed that anchoring training in real human data prevents the formation of brittle policies: using an Overcooked environment, they collected paired human trajectories and applied BC to construct a `human proxy' policy. Training an RL agent alongside this suboptimal partner forced it to account for arbitrary human conventions, delays and errors. This prepares agents for more effective ZSC, but does not address the volume of human data required to do so.

Although diverse agent pools have been established to improve performance in population-based training, ensuring that diversity presents further challenges.~\citet{lupu2021trajectory} addressed this with Trajectory Diversity (TrajeDi), adding an objective rewarding agents for behavioural difference from every other population member alongside the expected environmental reward; in Hanabi, agents best-responding to a TrajeDi population outperformed self-play agents under ZSC with novel partners.~\citet{strouse2021collaborating} translated such principles to human-AI teaming: their FCP framework achieved robust ZSC with live human partners in Overcooked without any human-in-the-loop training data, constructing a diverse population from historical checkpoints of self-play agents at various stages of learning.

The problem of coordinating with unknown teammates was formalised as ad hoc teamwork by~\citet{stone2010adhoc}: collaborating without prior opportunity to establish shared conventions.~\citet{hu2020other} demonstrated why self-play agents fail under these conditions, converging on arbitrary conventions a novel partner has no means of anticipating, and proposed Other-Play to avoid them.~\citet{laidlaw2022boltzmann} instead model the partner directly, characterising human decision-making as noisily rational and deriving a distribution over plausible policies to best-respond against. More recent population-based methods broaden the training distribution itself, either rewarding agents for developing policies incompatible with those of their peers~\citep{charakorn2023generating} or maximising the entropy of the population as a whole~\citep{zhao2023maximum}. \looseness = -1

Together these establish that populations diverse enough to prepare agents for ZSC can be generated without exhaustive human datasets or manual tuning. But where TrajeDi forces artificial trajectory divergence and FCP relies on the transient artefacts of under-trained networks, we pivot the source of population variance to procedural constraints: a training pool of partners subject to unique cognitive processing delays and perceptual limitations, generating trajectory diversity while ensuring the teaming agent adapts to a spectrum of psychologically valid, bounded behaviours.

\section{Partitioned Lunar Lander Environment}
\label{sec:environment}

\begin{figure}[htbp]
    \centering
    \includegraphics[width=0.44\linewidth]{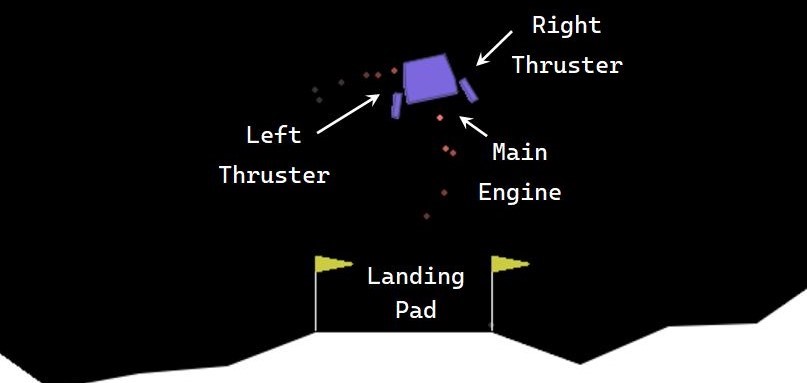}
    \hfill
    \includegraphics[width=0.44\linewidth]{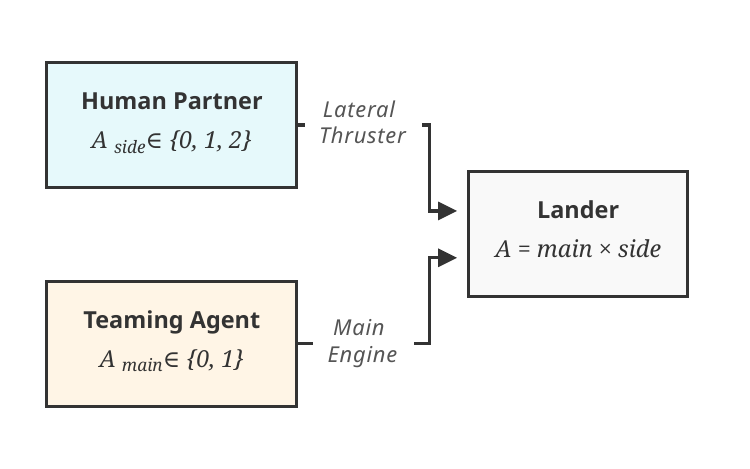}
    \caption{Lunar Lander environment from Gymnasium~\citep{gymnasium_lunarlander}. Physical geometry (left) and the partitioned control pipeline (right). The main engine and side thrusters must be coordinated to reach the landing pad.}
    \label{fig:lander}
\end{figure}

To address the gap in continuous interdependence identified above, we utilise a modified continuous Lunar Lander environment from Gymnasium~\citep{gymnasium_lunarlander}. The core objective requires landing safely within a designated target zone using a primary vertical thruster and two lateral thrusters (Figure~\ref{fig:lander}). Episodes initialise with random top-screen spawns and terminate via landing or crashing. This environment offers enough complexity to capture the nuance of continuous control demands found in real-world HMT tasks, yet remains computationally tractable enough to isolate the cooperative dynamics from the agent's baseline proficiency in the environment's physics. 

Unlike previous adoptions of this environment for teaming tasks, here the action space is decomposed. By assigning authority over the primary engine and lateral thrusters to distinct partners, the environment precludes unilateral task completion. This structural division mandates genuine real-time collaboration, requiring continuous coordination across the joint action space.

The state space is the standard 8-dimensional vector $\mathcal{S} = [x, y, \dot{x}, \dot{y}, \theta, \dot{\theta}, l, r]$, giving the coordinates, linear velocities, angle relative to the vertical axis, angular velocity, and contact indicators for the left and right legs. Neither of the environment's built-in control schemes suits split control. The discrete scheme offers four mutually exclusive actions, doing nothing or firing one of the left, main and right thrusters, so the two roles cannot act on the same frame. The continuous scheme takes upward and sideways forces on $[-1, 1]$, requiring precision no player with a keyboard can supply. To allow the two parties to act concurrently we implement a multi-discrete wrapper over the continuous environment, with $\mathcal{A} = \mathcal{A}_{\text{main}} \times \mathcal{A}_{\text{side}}$, where $\mathcal{A}_{\text{main}} = \{0, 1\}$ sets the main engine state (Off/On) and $\mathcal{A}_{\text{side}} = \{0, 1, 2\}$ the lateral thrusters (Off/Left/Right). Each discrete selection maps to a fixed magnitude in the underlying continuous space: the main engine fires at full power, while the lateral thrusters fire at 70\% of their maximum, full lateral thrust proving excessive for the scale of correction this task requires.

Both partners share a single reward, derived from the environment's shaping potential and charged per frame for each thruster in use, so that at each step the agent receives
\begin{equation}
  R_t = \Phi(s_t) - \Phi(s_{t-1}) - 0.3\,u_{\text{main},t} - 0.021\,u_{\text{side},t},
  \label{eq:reward}
\end{equation}
where $u_{\text{main},t}, u_{\text{side},t} \in \{0,1\}$ indicate whether each thruster is firing and $\Phi$ is that potential, which penalises distance from the pad, speed and tilt and rewards leg contact~\citep{gymnasium_lunarlander}. The lateral cost is below the nominal 0.03 because of this reduced power (Appendix~\ref{app:reward}). On the terminal frame the reward is replaced by $+100 - P(d)$ for a landing and $-100$ for a crash or for leaving the screen, with truncated episodes receiving neither. Since this reward permits landing anywhere on the terrain, whereas players naturally aim between the flags, a placement penalty realigns the two: with $d$ the horizontal distance from the centre of the landing pad, $P(d) = 0$ for $d \leq 0.2$ and $P(d) = 25 + 93.75(d - 0.2)$ otherwise, so landing just outside the zone costs 25 points, rising to 100 at the edge of the screen.

Reward shaping was considered for generating a diverse human-like population, forcing biases towards risk-taking, fuel economy or caution. Although this has improved teaming under ZSC with human partners~\citep{lupu2021trajectory, nazir2025zeroshot, powell2026zeroshot}, shaping explicitly to force diversity encodes designer bias, driving agents into artificial niches rather than capturing authentic human variance. We instead retain the standard incentives otherwise unaltered, so the constrained agents optimise the same objective as people. Diversity is located in computational rationality~\citep{gershman2015computational}, in which behaviour is utility maximisation under cognitive and physical limits. Observable constraints such as sensory latency and motor bottlenecks are also more empirically tractable than unobservable latent reward preferences. \looseness = -1

\section{Cognitively Constrained Partners}
\label{sec:partners}

Cognitive biases and heuristic thinking inherent to human reasoning have long been seen as evidence that human thinking is far from optimal~\citep{tversky1974judgment}, yet recent developments support the notion that human cognition is best understood as the optimal use of limited computational resources~\citep{lieder2020resource}. Building upon this, the constrained agents are designed to generate human-like behaviour by optimising within physiologically grounded constraints. Their policies should converge towards the hesitant, temporally chunked decision-making of human gameplay. To achieve this, a psychological wrapper is applied to the agent's environment which limits reaction speed, ability to perceive changes, and action frequency.

The agent's decision-making pipeline is deconstructed into three distinct phases: processing (thinking) time, mechanical delay, and a refractory (mental recovery) period. Since human reaction times consistently exhibit a positive skew due to attentional fluctuations, intra-episode delays for each phase are drawn from an exponentially modified Gaussian distribution, which serves as the standard mathematical model for reaction time variance within individuals~\citep{whelan2008effective}. To generate a diverse population of unique constrained agents, the mean ($\mu$) for each delay category is randomly initialised from a uniform distribution covering the spectrum of reaction speeds expected of real human players. The simulated environment runs at 50 frames per second, meaning that each frame of the game lasts 20\,ms. The processing delay, representing the central nervous system's perceptual and cognitive cycle, is therefore initialised with a mean uniformly between 4.0 and 5.5 frames in duration. This translates to an 80--110\,ms window, aligning with the Model Human Processor~\citep{card1983psychology}, which empirically establishes the human perceptual-cognitive cycle time at approximately 110\,ms. 

Similarly, the mechanical delay, representing the neuromuscular lag required to actuate a control input, is initialised with a mean between 3.0 and 4.0 frames which maps to the 70\,ms motor processor cycle defined in the same human-computer interaction framework~\citep{card1983psychology}. When combined, the processing and mechanical delays range from 140--190\,ms. This range of delays is also validated by the biological constraints observed in continuous human manual tracking and control tasks~\citep{McRuer1967Review}. Finally, following the execution of an action, the agent is subject to a refractory period sampled with a mean between 6.0 and 7.5 frames (120--150\,ms). This models the known psychological refractory period bottleneck observed when human players attempt to process sequential stimuli or rapidly switch tasks~\citep{allen1998psychological, strobach2012video}.

Across all three phases the standard deviation ($\sigma$) is held small (0.6/0.5/1.0), isolating low-level physiological jitter so that sensory-motor transmission remains tight. Distributional shape is instead governed by the exponential parameter ($\tau$). The processing and refractory phases carry a heavy tail ($\tau = 1.5$), modelling the vulnerability of complex decision-making and mental recovery to micro-distractions~\citep{unsworth2010lapses}, while the mechanical phase is near-symmetric, reflecting the reflex-driven nature of muscle actuation once a decision is finalised ($\tau = 0.5$). The distributions for all three phases, and for the convolved processing-plus-mechanical latency, are plotted in Appendix~\ref{app:delay}.

At the start of each episode, and whenever the agent commits to a new input, delays are drawn for each category. Releasing an input does not trigger this, reflecting the lower cognitive cost of ceasing an action relative to initiating one. Rather than receiving live observations, the agent is fed from a buffer in line with its current processing delay. Following the refractory period, the agent can perform an action, which is sent to the environment after its mechanical delay period. Actions must persist for at least four frames, reflecting the minimum human input duration of 80\,ms~\citep{oulasvirta2018neuromechanics}.

The delays make the problem partially observable, so a Recurrent PPO network~\citep{schulman2017proximal} was chosen over a typical feedforward approach. As well as mirroring human working memory, the hidden state within acts as a continuous cognitive buffer, allowing integration of sequential observations over time to bridge the temporal blind spots caused by the imposed reaction delays. Consequently, the agent generates temporally consistent inputs that replicate human goal-oriented behaviour, rather than reacting blindly to isolated frames. 

Beyond reaction delays, the wrapper restricts the agent's ability to detect small changes in velocity and orientation. While unconstrained agents register state changes instantaneously, human perception is bounded by Weber's law, which holds that the minimum detectable change in a stimulus is a constant proportion of that stimulus. This is true for human observation of both velocity~\citep{brown1960weber} and angular rotation~\citep{chen1996angle}. Three channels of the state vector are subject to this filter: the horizontal and vertical velocities, and the tilt angle. Each channel $q$ maintains its own record $q_{\mathrm{last}}$ of the value at which it was last registered. Where the difference falls short of the agent's threshold, $q_{\mathrm{last}}$ is held in place of the true value and the record left unchanged; where it meets the threshold, the change is registered and the record updated. Since comparison is against $q_{\mathrm{last}}$ rather than the preceding frame, a change too gradual to notice on any single frame accumulates until it crosses the threshold, so no quantity can drift arbitrarily far from what the agent believes it to be. The remaining channels pass through the processing delay unfiltered. 

A strict proportional rule cannot be applied directly: Weber's law breaks down at low stimulus intensities, where a constant fraction implies unbounded sensitivity and would grant the agent superhuman acuity during the precise landing phase. Each threshold is therefore bounded below, and the tilt threshold is further reduced near the cardinal angles, reflecting the greater sensitivity humans show to rotations close to vertical and horizontal, known as the `oblique effect'~\citep{furmanski2000oblique}. These departures leave the constrained agents less sensitive than a literal reading of Weber's law would make them, not more. The thresholds are given in full in Appendix~\ref{app:perception}.

To diversify the population, each agent is assigned a unique perceptual sensitivity before training. A single scalar $p \sim \mathcal{U}(0,1)$ is drawn and mapped linearly onto both Weber fractions, spanning the 4.5--8\% range reported for angle judgement~\citep{chen1996angle} and the 7--11\% range reported for velocity discrimination~\citep{brown1960weber}. Linking both thresholds to a shared variable mechanically couples the perceptual traits, ensuring that an agent with acute spatial awareness also possesses above-average velocity tracking, effectively simulating a generalised visual aptitude.

\section{Teaming Agent Implementation and Baselines}

\subsection{Teaming Agent}

Solo agents are first trained without the psychological wrapper to learn the environment's physics, while the cognitively constrained partners are trained alone from scratch on the full action space, each under an independently sampled wrapper with its own reaction-time and perceptual constraints. The task is then split: the constrained partner, its policy now fixed, contributes only its lateral output, while the teaming agent controls the main engine, its action space reduced to $\mathcal{A}_{\text{main}} = \{0, 1\}$ (Off/On). The teaming agent is initialised by partial weight transfer from a solo agent, so it begins already able to fly and spends its training budget learning to integrate its partner's behaviour. Since neither party observes the other's forthcoming decision, the partition adds a second source of partial observability alongside the partner's delays, and the recurrent architecture is retained for that reason. \looseness -1

At each time step the partner is queried for its lateral input, subject to the same delays, refractory period and minimum press duration under which it was trained, and this input, together with the teaming agent's main-engine action, steps the environment forward. The partner's \emph{executed} action $a_{\text{side}}$ (which, owing to the motor pipeline, is frequently not the one its policy requested) is appended as a ninth observation for the teaming agent's subsequent timestep. Each episode the teaming agent faces a partner drawn uniformly from a 45-agent pool, each with distinct delay and perceptual constraints. Appendices~\ref{app:hyperparameters} and~\ref{app:transfer} give the full hyperparameters and the transfer procedure. \looseness -1

\subsection{Baselines}
\label{sec:baselines}

Despite significant advancements in ZSC for HMT tasks, the RL literature predominantly focuses on tasks that are either symmetric in their action spaces or rely on discrete, turn-based mechanics where time is afforded before deciding the next action. In contrast, our strictly partitioned continuous control task demands real-time, reflex-driven coordination, representing a class of environments that remains significantly under-explored. Consequently, there is to our knowledge no single benchmark tailored for this specific problem domain, and we compare instead against three approaches adapted to this setting: FCP as the primary baseline, alongside PBT and standard MARL.

Adapting these methodologies to the partitioned control scheme of this environment necessitates a structural departure from their traditional implementations. First, concurrent learning is avoided wherever the comparison permits. Simultaneous gradient updates across the disjoint main and lateral action spaces induce severe environmental non-stationarity~\citep{lowe2017multi}, and, more consequentially for ZSC, mutually adapting agents converge on interdependent joint policies. This `secret handshake'~\citep{robson1990efficiency, hu2021off} is a convention a novel human partner does not share~\citep{strouse2021collaborating, carroll2019utility}. PBT and FCP therefore compute a Best Response against frozen partners, stabilising the transition dynamics into a stationary MDP and forcing the active agent to carry the burden of adaptation, as it must at deployment. MARL retains co-adaptation deliberately, since unconstrained joint optimisation is what that baseline exists to represent; it uses a staggered rather than simultaneous update schedule to control gradient instability.

To ensure equal learning opportunity across frameworks, all training pools are matched at 45 partners: 45 cognitively constrained partners, 15 FCP lineages sampled at three skill tiers, and a 46-agent PBT population in which each learner trains against the 45 others. Each condition is subsequently evaluated against a further 50 partners disjoint from their training pools. Matching these pool sizes ensures the comparison isolates the source of partner diversity rather than its quantity.

The information available to each partner necessarily differs between conditions. The MARL partner observes its teammate's action, as mutual co-adaptation requires it, whereas the partners used by FCP, PBT and our own method receive only the physical state (plus, for the constrained partners, an efference copy of their own input, Appendix~\ref{app:efference}). This distinction is deliberate: a partner given a noiseless per-frame feed of its teammate's action would be considerably more reactive than any person, and an agent best-responding to such a partner would come to depend on assistance a human will not provide. Implementation details for each baseline are given in Appendix~\ref{app:baselines}.

\section{Preliminary Results}

\subsection{Behavioural Validation}
\label{sec:behaviour}

This section provides a preliminary validation of the constrained partners against human play, based on a small informal pilot whose results are indicative rather than conclusive. Twenty participants from the authors' research department and their colleagues, all new to the game, each played 10 solo episodes on 10 fixed seeds, which were subsequently played by 45 cognitively constrained and 45 unconstrained agents. The unconstrained agents were measured once; since they are deterministic, repeated rollouts would produce identical behaviour given the use of fixed seeds. Conversely, the delays applied to the cognitively constrained agents are stochastic, so each agent repeated every scenario three times so as not to under-sample the ex-Gaussian distributions. Solo play was measured rather than cooperative play, as the absence of a teammate ensures that any observed difference is attributable to the controller alone.

Twenty-two features were extracted from each episode, characterising control behaviour as well as performance. They cover how long and how often each control is pressed and how variable those presses are, how input is divided between each of the thrusters, how quickly the controller responds to tilt, how the craft moves during descent, and the state in which the episode ends. These are all expressed as rates or durations rather than counts, to avoid confounding by episode length, and each is explained in Appendix~\ref{app:extended}. Episodes were averaged per controller, as those produced by the same controller constitute repeated sampling of a single policy rather than independent observations.

Divergence between controller populations was measured by energy distance~\citep{szekely2013energy}:

\begin{equation}
  \mathcal{E}(X, Y) = 2\,\mathrm{E}\lvert X - Y \rvert - \mathrm{E}\lvert X - X' \rvert - \mathrm{E}\lvert Y - Y' \rvert,
\end{equation}

where $X$ and $X'$ are two controllers drawn at random from one population, and $Y$ and $Y'$ from the other. The first term measures how far apart the two populations are, and the others subtract each population's internal spread, so a highly variable population is not mistaken for a distant one. $\mathcal{E}$ is 0 only for identical distributions and grows as they separate, so a lower distance from the human sample means an agent population whose play is harder to distinguish from that of real players. As $\mathcal{E}$ has no absolute scale, each value is read against the distance between two random halves of the human sample. Over 2,000 random splits a distance of 1.19 (95\% CI $[0.57, 3.03]$) was found between the two human populations, so a genuinely indistinguishable population would be expected to fall within this interval. \looseness -1

Following feature standardisation, human players lay 4.12 from the constrained agents and 7.41 from the unconstrained, a difference of $+3.29$ (95\% CI $[+1.64, +4.86]$) over 5,000 bootstrap resamples of controllers, with the constrained agents closer in 99.9\% of samples. The constrained agents therefore lie roughly 3.5 times further from human players than human players lie from one another, against 6.2 times for the unconstrained agents. Since the human sample is small, with only 10 players on each side of a split, the human-to-human distance varied widely across the 2,000 resamples, which limits how precisely these distances can be estimated; the result therefore supports the ordering of the two agent populations relative to human players more firmly than any claim about how close either lies to them. \looseness -1

Energy distance shows whether populations differ but not where, so three measures are also reported per feature: the difference in group means, the Wasserstein distance normalised by group spread (lower is closer), and the overlap in probability mass (higher is closer), with definitions and limitations given in Appendix~\ref{app:extended}. The constrained agents lie closer to human play on 15 of the 22 features under each measure independently, and the three measures agree on 17 (Tables~\ref{tab:features} and~\ref{tab:raw_features}).

\begin{figure}[!htbp]
    \centering
    \includegraphics[width=0.85\linewidth]{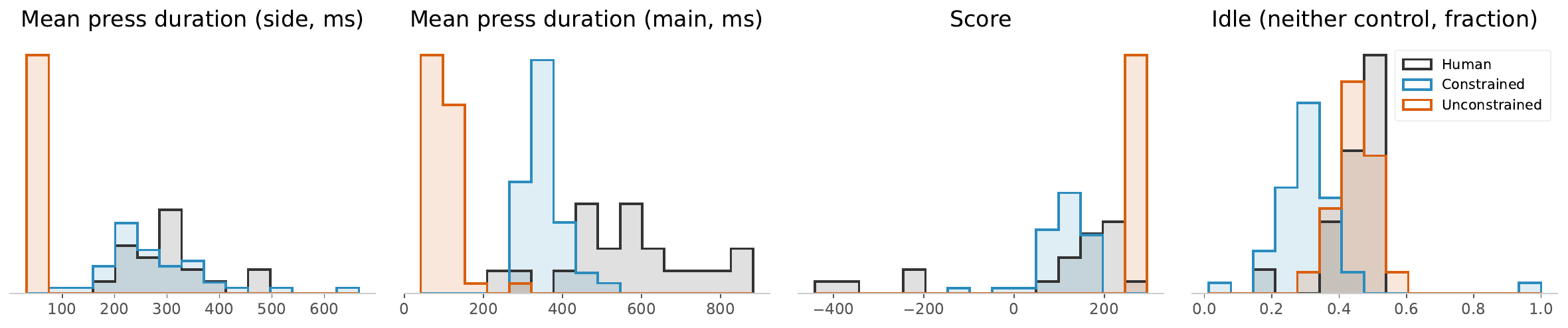}
    \caption{Individual controller distributions for mean press duration (side and main), score, and idle time (neither control) across human, constrained, and unconstrained groups.}
    \label{fig:features_core}
\end{figure}

Mean press duration for the side thrusters is the closest match of any feature, sharing 73\% of its probability mass with participants where the unconstrained agents share none, so the constraints reproduce human input lengths almost exactly (Figure~\ref{fig:features_core}). The main-engine presses are closer than the unconstrained agents manage but still short, at 353\,ms against 98\,ms, with participants averaging 562\,ms. Participants also vary widely in how long they hold the main engine (SD 171\,ms) where the constrained agents cluster tightly (SD 49\,ms), so the constraint lengthens presses towards the human range without reproducing the variation between players.

Participants pressed the lateral thrusters 0.94 times per second and the constrained agents 1.62, against 6.12 for the unconstrained agents; for bursts of three or more direction changes within half a second the rates are 0.43, 1.39 and 9.73. No player could sustain the rate the unconstrained agents produce, and the constraints bring the agents to within a small multiple of human behaviour where they would otherwise be an order of magnitude adrift. All four terminal measures (displacement from the centre, speed, tilt and score) also favour the constrained agents, with terminal tilt the highest overlap of any feature at 78\%, although no constraint pertains to how an episode should end.

Five features favour the unconstrained agents under all three measures, all concerning idle time, directional preference, or tilt-correction latency. The unconstrained agents are almost indistinguishable from participants in the proportion of frames with the side thrusters inactive (0.714 against 0.696, versus 0.565 for constrained) and with neither control active (0.446 against 0.452, versus 0.311). Participants also show a  directional bias in their lateral input (0.34) which the unconstrained agents partially reproduce (0.17) and the constrained do not (0.10), and correct tilt more slowly (286\,ms) than either the constrained (131\,ms) or unconstrained (146\,ms) agents, the latter marginally closer.

\subsection{Cross-Play Generalisation}
\label{sec:crossplay}

\begin{figure}[!htbp]
    \centering
    \includegraphics[width=0.6\linewidth]{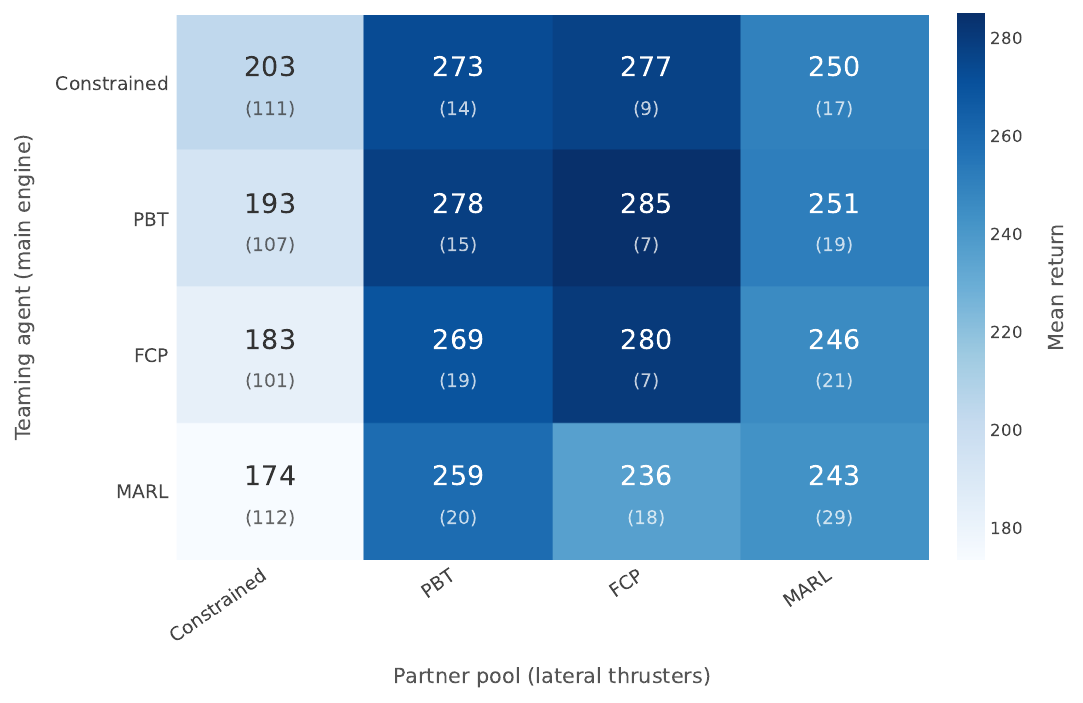}
    \caption{Mean return of each agent–partner pairing, averaged over 10 independently trained replicates per agent type. Parenthesised figures are the standard deviation across the 15 partners comprising each pool. One of the fifteen MARL side agents is the co-adapted partner of each MARL replicate.}
    \label{fig:crossplay}
\end{figure}

Partners are grouped into four families by the procedure that generated them (constrained, PBT, FCP and MARL), and each teaming agent's matched family is the one it trained against, the other three being held out. Figure~\ref{fig:crossplay} shows the mean return for every agent against every family. Each condition is represented by ten independently trained replicates rather than a single agent, since one training run conflates a method's behaviour with the outcome of that particular run; every replicate plays fifteen partners from each family on twenty fixed seeds, twice, giving 6{,}000 episodes per cell and 96{,}000 in total. Every partner in every pool is drawn from an evaluation set disjoint from the training pools of all four methods, so no agent has met any of these partners before, bar the exception noted in Figure~\ref{fig:crossplay}. The diagonal therefore does not pair an agent with partners it trained against, but with unseen partners produced by the same generative procedure; the off-diagonal cells pair it with partners produced by a procedure it has never encountered, which is the situation a human teammate presents.

The clearest result is what happens when an agent moves away from the procedure it was trained against. Table~\ref{tab:ownpool} reports each agent's return against its matched pool and against the three it did not train alongside. The constrained agent is the only one to improve, rising from 203.4 to 266.6; PBT falls by 34.9 points and FCP by 47.6, both intervals excluding zero, while MARL's smaller decline of 19.9 is not distinguishable from zero. Because the comparison is paired by replicate, the strength of any individual training run cancels from the difference rather than adding noise to it. PBT and FCP agents thus depend on the procedure that produced their partners, in a way that agents trained against cognitively constrained partners do not.

\begin{table}[!ht]
\centering
\caption{Mean return when each agent is paired with partners from the pool generated by the procedure it trained against and from the three pools it did not train against. Intervals are percentiles of a two-way cluster bootstrap over 5{,}000 resamples of trained replicates and partners, paired by replicate so that the strength of a given training run cancels from the difference.}
\label{tab:ownpool}
\begin{tabular}{lcccc}
\toprule
Teaming agent & Matched pool & Held out & Change & 95\% CI \\
\midrule
Constrained & 203.4 & 266.6 & $+63.1$ & [$+23.1, +125.3$] \\
PBT & 277.9 & 242.9 & $-34.9$ & [$-58.3, -15.8$] \\
FCP & 280.1 & 232.6 & $-47.6$ & [$-70.7, -28.9$] \\
MARL & 242.7 & 222.7 & $-19.9$ & [$-51.1, +5.5$] \\
\bottomrule
\end{tabular}
\end{table}

That property is not accompanied by an advantage in raw performance. Restricting the between-agent comparison to families neither agent trained alongside, so that both are averaged over exactly the same thirty partners, the constrained agent is indistinguishable from PBT ($-4.5$, 95\% CI $[-11.4, +1.9]$) and from FCP ($+4.0$, $[-1.8, +10.0]$), and ahead of MARL ($+27.5$, $[+8.7, +52.7]$; Table~\ref{tab:heldout}). Averaged across all four pools the three population-based methods are close, at 250.8 for the constrained agent, 251.7 for PBT and 244.4 for FCP, with MARL at 227.7. Training against delayed and inconsistent partners therefore costs nothing against competent ones.

Every agent performs worst against the constrained pool, which is also far more heterogeneous than the others: the standard deviation across its fifteen members exceeds 100 return points for every agent, where the FCP, PBT and MARL pools fall between 7 and 29. Delayed and perceptually limited partners therefore differ from one another to a degree that conventionally generated populations do not reproduce. Against it the constrained agent returns 203.4, ahead of FCP by 20.8 ($[+0.1, +43.7]$) and MARL by 29.9 ($[+13.0, +50.7]$), though not distinguishable from PBT ($+10.5$, $[-7.1, +30.5]$). These partners are unseen by every agent, but only the constrained agent trained against the procedure that generated them, so the comparison should be read with that asymmetry in mind.

\section{Human Study Design}
\label{sec:study}

Cross-play against synthetic partners is the field's standard proxy for teaming performance, but how well its rankings predict coordination with real people is rarely tested directly. To measure the efficacy of the proposed architecture, and to test whether the cross-play rankings of Section~\ref{sec:crossplay} persist with human partners, a formal within-subject study that has received ethical approval from the authors' institution is scheduled to take place. This study will see 100 participants play the game alone, and with each of the four teaming agent types (ten instances of each). Scores will be analysed using a linear mixed-effects model, with the agent as a fixed effect and the participant as a random intercept, so that agents are compared relative to each individual's own baseline rather than to the group mean. Planned contrasts then compare the proposed agent against each baseline in turn. Full details for the planned study, model and the secondary analyses are set out in Appendix~\ref{app:analysis}.

\section{Limitations}

Humans are not static policies themselves, and adapt to their partners over time; our work focuses specifically on the first instances of cooperation, in which a player approaches the task with a relatively fixed set of cognitive heuristics and physiological constraints. This architecture is intended as a strong starting point for more specialised human-in-the-loop training, reducing the effort needed to reach an agent policy which acts as a best response to novel human partners.

The cross-play evaluation reported in Section~\ref{sec:crossplay} pairs teaming agents with held-out artificial partners, and consequently demonstrates generalisation across partner types rather than to human players specifically. Although the behavioural validation partially bridges this gap by establishing that our partners occupy a similar behavioural region to real players, the inference from cross-play performance to human teaming performance remains untested until the study is complete.

The constrained pool is the hardest of the four for every agent, so the constrained agent begins from the lowest matched-pool baseline and an improvement is easier to obtain from 203.4 than from PBT's 277.9 or FCP's 280.1. The change reported in Table~\ref{tab:ownpool} is therefore not directly comparable across agents. The comparison in Table~\ref{tab:heldout}, where every agent meets the same partners, is free of this effect, and shows parity with PBT and FCP rather than an advantage over them.

The constrained partners perceive only the physical state, whereas a player would also see the main engine's exhaust plume. As set out in Section~\ref{sec:baselines}, this under-models partner awareness deliberately, yielding a cautious policy more analogous to a player concentrating on their own role. A more faithful treatment would grant the partner intermittent awareness of its teammate's actions, filtered through the same perceptual pipeline as the rest of its observations.

Although restrictions on the constrained agents' perception of changes in both speed and angle are imposed, these do not cover all channels pertaining to those variables, and it is possible that the agent infers what it has not been allowed to see. Angular velocity in particular passes through unfiltered, so an agent whose tilt perception is hindered still receives a clean signal of the rate at which that tilt is changing, from which the true angle can in principle be integrated. Extending these restrictions to every channel, granting partners filtered awareness of their teammate, and building more specialised human-in-the-loop training on this starting point are all left to future work.

\section{Conclusion}

This paper has presented a framework in which partner diversity for ZSC is generated procedurally from perceptual and motor constraints grounded in human physiological literature, applied to a partitioned environment in which neither party can complete the objective alone. The framework's central claim, that agents trained against these partners will coordinate more effectively with human players, awaits the study described in Section~\ref{sec:study}, but the preliminary results establish the two conditions it depends upon: the partner population occupies a similar region of behavioural space to real players, though without reproducing the full variation between them, and agents trained against it match the strongest baselines under cross-play while losing nothing when the partner-generation procedure changes, where PBT and FCP degrade. Beyond ranking the agents, the study will show whether the orderings produced by cross-play evaluation survive contact with human partners, a direct measure of how well this standard proxy predicts performance in real human-agent teams.

\newpage
\bibliographystyle{plainnat}
\bibliography{bibliography}

\newpage
\appendix
\raggedbottom

\section{Implementation Details}
\label{app:implementation}

\subsection{Reward Function}
\label{app:reward}

As in the standard Lunar Lander environment~\citep{gymnasium_lunarlander}, the continuous step-wise reward in Equation~\eqref{eq:reward} is derived from the difference in potential between consecutive states, incentivising the agent to continuously minimise its speed, tilt, and distance from the goal. Per-frame penalties are applied for fuel consumption to discourage inefficient hovering or erratic manoeuvres. The environment charges these penalties in proportion to the power at which each thruster fires, so the fixed magnitudes imposed by the multi-discrete wrapper of Section~\ref{sec:environment} determine the effective coefficients: the main engine, firing at full power, incurs the nominal $0.3$ per frame, while the lateral thrusters, firing at $0.7$, incur $0.021$ rather than the typical $0.03$. Lateral input is therefore roughly fourteen times cheaper per frame than main-engine input, as opposed to the environment's ten-to-one ratio at full side thruster power, though the weaker thrust means a given correction must be held for longer. The final step's reward is replaced by $+100$ for a landing or $-100$ for a crash, including leaving the screen, and $P(d)$ is deducted from landings only.

\subsection{Ex-Gaussian Delay Distribution}
\label{app:delay}

\begin{figure}[htbp]
    \centering
    \includegraphics[width=0.95\textwidth]{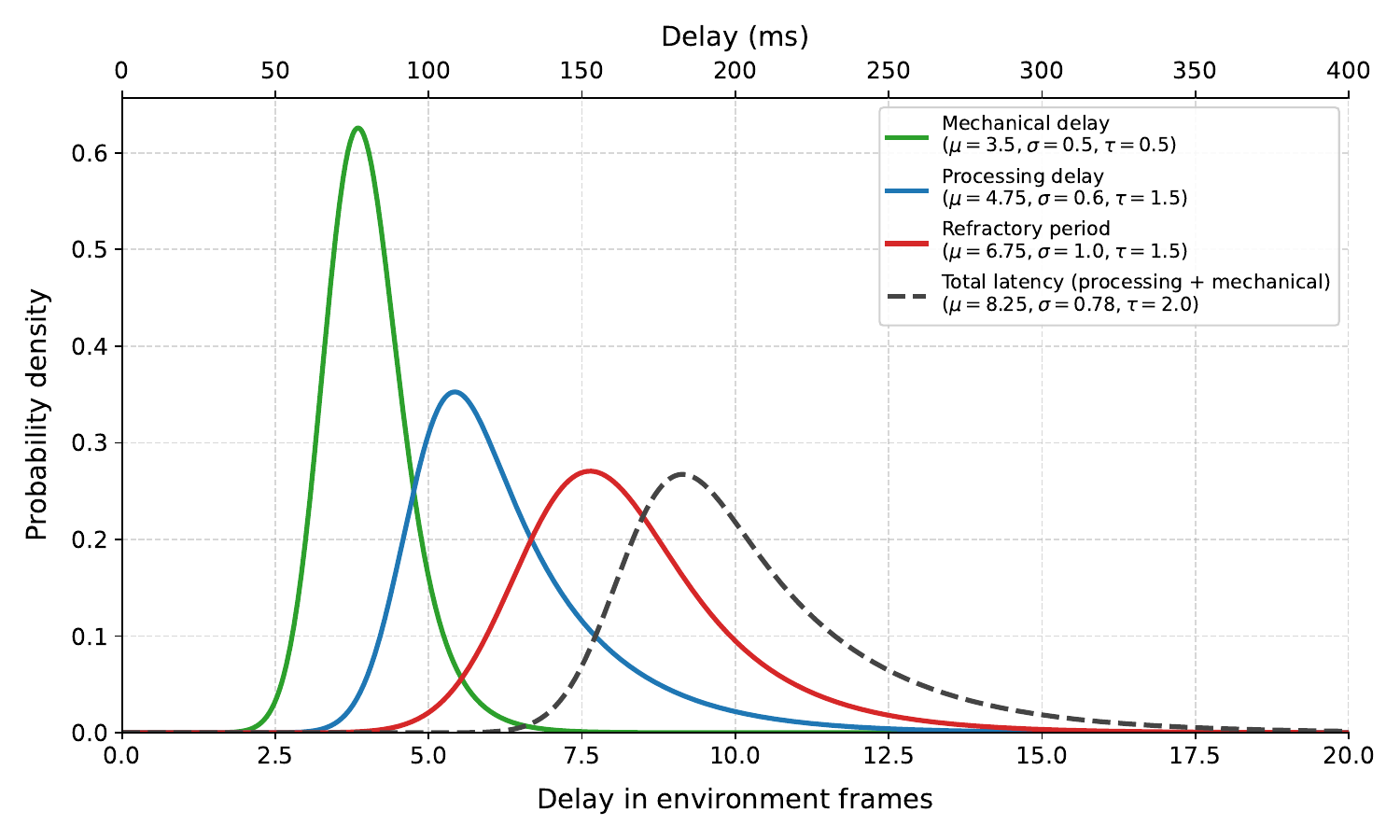}
    \caption{Ex-Gaussian delay distributions for a single constrained agent, shown at the midpoint of each parameter range. The mean of every phase is sampled independently for each constrained agent, so each member of the population carries a different set of curves. The dashed curve is the convolution of the processing and mechanical delays, giving the total latency between a change in the environment and the corresponding change in the partner's output.}
    \label{fig:ex_gaussian_delays}
\end{figure}

\subsection{Perceptual Thresholds}
\label{app:perception}

Section~\ref{sec:partners} describes a perceptual filter applied to the horizontal and vertical velocities and to the tilt angle, with each channel compared against the value at which it was last registered. The agent's two Weber fractions are drawn before training from a single scalar $p \sim \mathcal{U}(0,1)$, mapped linearly onto the ranges reported in the psychophysical literature:

\begin{equation}
  w_\theta = 0.045 + 0.035\,p,
  \qquad
  w_v = 0.07 + 0.04\,p .
\end{equation}

For the velocities the threshold is $\max\!\left(0.02,\; w_v \lvert \dot{q}_{\mathrm{last}} \rvert\right)$, applied separately to $\dot{x}$ and $\dot{y}$ with $\dot{q}$ denoting whichever is being tested. The Weber fraction governs where it exceeds the floor, and the constant otherwise; the crossover falls at roughly $\lvert \dot{q}_{\mathrm{last}} \rvert = 0.2$--$0.3$, so the floor is the operative term through a substantial portion of a controlled descent.

For tilt the correction takes the form of a constant offset rather than a floor, giving a threshold of $w_\theta \left( \lvert \theta_{\mathrm{last}} \rvert + 1 \right)$. The offset sets a baseline below which no change is perceptible regardless of orientation, so the agent requires a change of $w_\theta$ radians when level and proportionally more as the craft tilts further. Since $\lvert \theta \rvert$ seldom exceeds $0.5$~rad, this yields a near-constant threshold of approximately 2.6--4.6$^{\circ}$, several times stricter than the ratio alone would give at these magnitudes.

An exception is the noted increased sensitivity in humans to rotational changes near the horizontal and vertical axes (cardinal angles), known as the `oblique effect'~\citep{furmanski2000oblique}. To replicate this, the tilt threshold is halved when $\lvert \theta_{\mathrm{last}} \rvert$ lies within $10^{\circ}$ of a cardinal angle and quartered when within $2^{\circ}$, so an agent flying close to level detects tilt changes two to four times smaller than one flying obliquely. The adjustment is keyed to the last perceived angle rather than the true one, since a player's acuity depends on where they believe the craft to be pointing. Since the lander is near-vertical for much of a controlled descent, the reduced threshold is the operative one throughout the approach, with the unmodified value applying mainly during recovery from a heavy tilt. Even under the strongest reduction the tilt threshold remains several times larger than a pure ratio would give at typical tilts.

The filter is applied to the buffered observation after the processing delay has been imposed. Both the difference and the threshold are evaluated on magnitudes, so a quantity passing through zero always registers. Each channel maintains its own record, so a registered change on one does not refresh the others. 

\subsection{Efference Copy}
\label{app:efference}

A player operating a keyboard is continuously aware of which key they are holding~-- this sensation is tactile and immediate, rather than inferred through observing what happens in the game. The constrained agent, by contrast, emits an intent each frame and receives no indication of whether that intent was committed, queued behind the mechanical delay, or discarded during a refractory period. To get this information, it would have to reconstruct its own motor state from delayed and filtered physics in a decidedly inhuman manner. To remove this deficit the agent observes the lateral input it is currently executing, expanding its observation space from eight dimensions to nine. This value is not subject to any delays or to the perceptual filter of Appendix~\ref{app:perception}, as efference copy (the internal copy of motor commands created by the brain) and tactile feedback are substantially faster than the visual pathway. The internal timers governing the refractory and commit windows are deliberately withheld, as a person has no introspective access to their own psychological refractory period.

\subsection{Baselines}
\label{app:baselines}

\subsubsection{MARL}
\label{app:marl}

Our standard MARL baseline represents traditional, unconstrained joint-optimisation, where agents co-adapt from scratch without pre-existing understanding of the underlying physics environment. However, applying simultaneous policy updates across disjoint continuous action spaces induces severe non-stationarity, rendering the environment's transition dynamics highly unstable from either agent's perspective. To resolve this ``moving-target'' problem while preserving the capacity for mutual adaptation, we implement a staggered training paradigm. By temporarily freezing one partner's policy, the transition dynamics are stabilised from the perspective of the active agent, effectively reducing the multi-agent challenge to a stationary, single-agent Markov Decision Process (MDP). This alternating approach circumvents the gradient instability inherent to simultaneous independent learning~\citep{hernandez2019survey}. Accordingly, our active learning role alternates between the main and lateral agents every 20,000 timesteps.

This interval equates approximately to 85 full episodes, and is intended to balance statistical significance with policy entropy. Extremely short intervals allow gradient updates to be dominated by the variance of random environmental initialisations, effectively recreating the moving-target instability. Conversely, prolonged active phases cause the agent to fully converge against its static partner, leading to catastrophic forgetting and highly brittle, exploitative policies~\citep{lanctot2017unified}. This overfitting can be prevented by truncating the learning process to compute an approximate best response~\citep{muller2019generalised}. This ensures the policy shifts enough to meaningfully improve coordination, without fully overfitting to the frozen partner, enabling stable co-adaptation.

To prevent swap intervals from terminating active simulations mid-episode, the agents train in completely decoupled environment instances. The active agent interacts with its dedicated environment while querying its frozen partner, ensuring experience collection spans uninterrupted episodes to avoid injecting bias into the policy's advantage estimation~\citep{gronauer2022multi}. Furthermore, this strictly partitioned control induces a Decentralised Partially Observable Markov Decision Process (Dec-POMDP). To mitigate this partial observability problem, each agent's observation space is appended with its partner's previously executed action in order to construct a basic action-observation history, a mathematically established mechanism for facilitating coordination in decentralised tasks~\citep{oliehoek2016concise}.

\subsubsection{PBT}
\label{app:pbt}

Before any multi-agent coordination begins, models are initially trained to develop a foundational understanding of the environment's physics. A random subset of these agents are then initialised using these pre-adapted weights, isolating the subsequent learning phase strictly to the teaming dynamics. During this collaborative phase, control is again partitioned such that the primary agent has control over the main engine, while the partner agent uses the lateral thrusters.

At the start of each episode, the environment randomly selects a frozen partner from the population of partner agents. The primary agent has no explicit knowledge of which static policy is controlling the lateral thrusters in a given episode, meaning this training effectively results in a generalised best response to the entire distribution of partners. As with the MARL framework, the partner agent's current input is appended to the primary agent's state space, although in this case the knowledge is shared unilaterally.

\subsubsection{FCP}
\label{app:fcp}

FCP computes a best response across a distribution of partners drawn from different stages of their learning trajectories, conventionally by checkpointing symmetric self-play agents~\citep{strouse2021collaborating}. Partitioned control requires two adaptations.

The first concerns pool generation. Training both roles from scratch simultaneously yields unrecoverable catastrophic failures, while initialising them from a pre-trained joint policy artificially flatlines the learning curve, precluding the extraction of intermediate skill checkpoints. We instead generate partners by computing an asymmetric best response against fixed unconstrained policies: a set of pre-trained, unconstrained main agents is frozen to provide the vertical stability necessary to prevent premature episode termination, and independent lateral agents are initialised from scratch and trained against these static main controllers. This stabilisation permits the lateral agents to generate the clean, upward-trending learning trajectories necessary for historical checkpointing.

The second concerns checkpoint selection. The standard framework samples from precisely the start, middle and end of training to maximise behavioural diversity. This is ill-suited to strictly partitioned continuous control: where a proficient agent in a symmetric task can compensate for a severely inept partner, extreme incompetence on the lateral thrusters produces repeated unrecoverable failures and no actionable learning signal. To ensure both population diversity and usable feedback, we sample at the 33\%, 66\% and 100\% marks of peak learning. Fifteen independently trained lineages sampled at three tiers give the 45-partner pool used in training.

\section{Training Details}
\label{app:training}

Experiments were run under Python 3.9.6 with Gymnasium 1.1.1, Stable-Baselines3 2.7.1~\citep{raffin2021stable} and sb3-contrib 2.7.1, PyTorch 2.8.0 and NumPy 2.0.2. All parameters not listed in Table~\ref{tab:hyperparameters} take the Stable-Baselines3 defaults for \texttt{RecurrentPPO} with an \texttt{MlpLstmPolicy}. Training uses a single environment instance, and no observation or reward normalisation is applied. Only the evaluation seeds are fixed; training runs are unseeded, so exact numerical reproduction of a given agent is not possible, although each constrained partner's sampled constraint profile is written to disk alongside its weights so that the population itself is recoverable.

All training was performed on CPU with a single PyTorch thread per run, with runs executed 32-wide across pinned physical cores. Training runs required up to 4 GB of memory each, and had an approximate duration of 8--12 hours per training instance.  No GPU was used. Each agent was trained for a maximum of 2,500,000 timesteps, with early stopping applied within that budget as described in Appendix~\ref{app:earlystopping}. The full project required considerably more compute than the runs reported here, as the training pipeline and constraint implementation were revised over several iterations whose agents are not reported.

\subsection{Hyperparameters}
\label{app:hyperparameters}

 All agents, across both the proposed framework and the baselines, use Recurrent Proximal Policy Optimisation with an \texttt{MlpLstmPolicy}, for the reasons given in Section~\ref{sec:partners}; in the teaming phase the recurrence additionally compensates for the partitioned action space, in which neither party observes the other's forthcoming decision.

Hyperparameters were established through a combination of standard convention and empirical testing during the training of the solo agents, and are identical across all conditions except where noted in Table~\ref{tab:hyperparameters}.

\begin{table}[!ht]
\centering
\caption{Training hyperparameters. All agents use RecurrentPPO with an
\texttt{MlpLstmPolicy}.}
\label{tab:hyperparameters}
\begin{tabular}{@{}lll@{}}
\toprule
Parameter & Value & Notes \\
\midrule
Discount factor ($\gamma$)    & 0.99      & Standard for continuous control \\
Learning rate               & 0.0005    & Constant throughout training \\
Entropy coefficient         & 0.005    & Balances exploration with destabilisation risk \\
Batch size                  & 128       & Empirically tuned on solo agents \\
Rollout length              & 1024      & Empirically tuned on solo agents \\
Episode limit               & 750 steps & Same for all conditions \\
Training budget             & 2{,}500{,}000 steps & 5{,}000{,}000 for MARL, split across two learners \\
Early-stopping patience     & 5{,}000 episodes & 50 evaluations at 100-episode intervals \\
Evaluation frequency         & 100 episodes & On 50 fixed deterministic seeds \\
Partner pool size           & 45        & Matched across non-MARL conditions \\
Evaluation partners         & 50        & Disjoint from the training pool \\
\bottomrule
\end{tabular}
\end{table}

\subsection{Weight Transfer}
\label{app:transfer}

The PBT, FCP and primary teaming agents are initialised from solo agents pre-trained on the environment's physics without a partner. Because the partner's executed lateral action is appended to the observation, the input layer expands from eight to nine dimensions. The LSTM and shared feature layers are transferred directly. Due to this expansion, the recurrent input matrix expands from eight to nine columns; the eight physical-state columns are copied and the ninth is randomly initialised. The action head is not transferred, since the solo agent's MultiDiscrete head does not match the teaming agent's binary main-engine head, so the agent inherits a learned representation of the environment's physics and relearns the output mapping over it. The MARL agents receive no transfer, since unconstrained joint optimisation from scratch is what that baseline represents. The cognitively constrained partners are similarly trained from scratch.

\subsection{Evaluation Protocol}

Evaluation is conducted every 100 training episodes against a fixed array of 50 deterministic seeds, with matched fixed partners in the teaming conditions. Holding both the environment seeds and the partner constant isolates policy improvement from variation in initial conditions or partner sampling.

The size of the evaluation set was calibrated to control false positives during checkpoint selection. Since the environment's returns are high-variance and bimodal, a checkpoint evaluated across too few seeds may post a higher mean reward through favourable initial conditions alone, causing it to be erroneously saved as the new best model and to reset the early-stopping counter. Evaluating against a very large seed pool would suppress this effect, but is computationally prohibitive given the number of checkpoints and agents involved.

A selection of agents was therefore evaluated at every training checkpoint across a pool of 100 fixed seeds, retaining the per-seed returns. For each pair of successive checkpoints exhibiting an increase in mean return, a paired one-tailed $t$-test was applied to progressively larger subsets of the seed pool in increments of ten. This established the subset size at which improvements of the magnitude typically observed during training were distinguishable from seed variability at the 5\% level. Fifty seeds proved sufficient across the agents and improvements examined, so this was adopted as the evaluation set size.

\subsection{Early Stopping}
\label{app:earlystopping}
Models were trained for a maximum of 2,500,000 timesteps, with the exception of the MARL condition, whose budget of 5,000,000 is divided between two concurrently trained agents so that each receives an equivalent 2,500,000.

To preserve computational resources while preventing premature termination, an early-stopping patience mechanism was implemented. Preliminary training visualisations of agents revealed that policies frequently underwent periods of severe temporary destabilisation, sometimes lasting several thousand episodes, before recovering to achieve a higher global optimum. A tolerant patience threshold was therefore adopted. Training terminates only when the mean evaluation reward fails to improve across 5,000 consecutive episodes, equivalent to 50 successive evaluations. This affords agents sufficient exploratory bandwidth to recover from performance dips during late-stage fine tuning.

\section{Future Analysis Plan}
\label{app:analysis}

\subsection{Human Study Details}
\label{app:study}

Initially, participants will play through 20 episodes of the game on their own to familiarise themselves with the mechanics and goals, ensuring a basic competence before the teaming phase begins. Participants will then play 20 episodes with each of the four teaming agent types (MARL, PBT, FCP, and the Cognitively Constrained Partner agent), 80 in total, with agent order randomised to mitigate ordering effects and learning bias. Finally, the participants will play 20 episodes alone again before completing a short debrief questionnaire. With each episode lasting roughly ten seconds, the full 120-episode protocol amounts to approximately 20 minutes of play, and participants may take breaks between episodes whenever they wish. Each teaming agent is colour-coded so that participants can retrospectively rank their preferences, and the questionnaire asks whether they felt each agent adapted to their intent and whether it made the game easier than playing alone. One hundred participants will take part in the study, which has been given ethical approval. They will be recruited through the university, with no restriction on gaming aptitude so as to ensure a broad spectrum of behaviour types. Since the design is within-subject, each agent contrast is estimated within participants rather than between them, and the 20 episodes played alongside each agent average out the episode-level variance the environment's bimodal returns produce.

Ten teaming agents will be trained under each condition for use in the study, and each of the four resulting lists of 10 agents will then be randomly shuffled separately. In order to ensure we can separate differences that could arise from different agent initialisations, the indexes of these agents in their respective lists will act to pair them. The 100 participants in the study will randomly be assigned an index at the start of the experiment such that each batch of agents is played against 10 times.

\subsection{Mixed-effects Model}
\label{app:mixed}

Let $y_{ij}$ denote the mean score achieved by participant $i$ when partnered with agent $j$, where $j$ indexes the four teaming conditions (Constrained, PBT, FCP, MARL). Each participant is assigned, at the start of the study, a single index $k_i \in \{1, \dots, 10\}$ drawn from a balanced random allocation, so that every value of $k_i$ is held by exactly ten participants. This index selects which of the ten agent instances retained for each condition (Section~\ref{sec:study}) the participant plays: participant $i$ is partnered under condition $j$ with the instance identified by $(j, k_i)$. Scores are modelled as

\[
y_{ij} = \beta_0 + \beta_j + u_i + v_{j,k_i} + \varepsilon_{ij}, \qquad
u_i \sim \mathcal{N}(0, \sigma_u^2), \quad
v_{j,k} \sim \mathcal{N}(0, \sigma_v^2), \quad
\varepsilon_{ij} \sim \mathcal{N}(0, \sigma^2),
\]

where $\beta_0$ is the expected score of a typical participant partnered with a typical instance of the reference agent, $\beta_j$ the change in expected score attributable to condition $j$, $u_i$ the baseline proficiency of participant $i$, $v_{j,k}$ the deviation of agent instance $k$ under condition $j$ from that condition's mean, and $\varepsilon_{ij}$ the residual variation unexplained by other terms. The constrained-partner agent is taken as the reference level, so each $\beta_j$ expresses a baseline's performance relative to it and the three planned contrasts are read directly from the fitted coefficients. Trial index is included as a covariate, since the order in which agents are encountered is randomised rather than balanced and participants may improve over the course of a
session.

Each participant is paired with every agent, so their scores are not independent: a proficient player scores highly under every condition and a struggling one poorly. Rather than fitting a separate baseline for each of the hundred participants, the model estimates the variance $\sigma_u^2$ of baseline proficiency across the population. This is recoverable from the data because participants who are more able score highly under every condition, so their four scores rise and fall together; the stronger that tendency, the more participants must differ from one another. Estimating it allows the variation between participants to be set aside, so that agents are compared within individuals rather than across them. Pilot data indicated substantial between-participant variation, with mean scores spanning several hundred points, which would otherwise be attributed to residual error and obscure any difference between agents.

The random effect $v_{j,k}$ serves a similar purpose for agent instances rather than participants. Ten agents are trained under each condition, so any single instance's result is one draw from the outcomes that condition's training procedure can produce, not a deterministic property of the condition itself. Without $v_{j,k}$, systematic differences between a condition's own instances (one agent initialisation that happened to train unusually well or poorly) would be absorbed into $\varepsilon_{ij}$ and reported as if they reflected the participant's response to the method in general. Because the balanced index assignment guarantees exactly ten participants per instance, $\sigma_v^2$ is identifiable within each condition from the spread of those ten instance means around the condition mean, and $\beta_j$ is correspondingly interpreted as the effect of a condition averaged over the instances it can produce, with the standard errors of the planned contrasts widened to reflect that averaging rather than treating all ten instances of a condition as interchangeable repetitions of one true effect.

Both $u_i$ and $v_{j,k}$ will be checked for normality after fitting. Pilot data suggests a bimodality in baseline proficiency, with participants tending either to land consistently or rarely succeed, so it is possible that the participant random intercepts depart from normality. Should this occur, $u_i$ will be modelled as a mixture of two normal components rather than a single one, corresponding to the two levels of competence apparent in the pilot. No comparable prior evidence exists for $v_{j,k}$, so it will be assumed normal.

A separate concern applies to $v_{j,k}$ regardless of its shape: with only ten instances per condition, its variance $\sigma_v^2$ will necessarily be estimated imprecisely, and diagnostics may be unable to distinguish it from zero. This does not threaten the model's main results; the fixed-effect estimates $\beta_j$ and their standard errors come from the same fit that includes $v_{j,k}$, and already reflect whatever instance-level variation is present regardless of how precisely $\sigma_v^2$ itself can be determined, so they remain the paper's primary result under either outcome. What changes is only whether $\sigma_v^2$ is reported as a finding in its own right, as if well-estimated, it will be presented alongside $\sigma_u^2$ as evidence of how much a participant's score depended on the specific trained instance they played. If it cannot be reliably distinguished from zero, no claim will be made about its magnitude beyond noting that it was included in, and accounted for by, the model.

This retains the structure of the model and the interpretation of the agent effects, while representing both the population of participants and the population of trained instances as they are rather than as single homogeneous groups.

\subsection{Secondary Analyses}
\label{app:secondary_analysis}

Since each participant plays multiple episodes alongside every agent, their scores cannot be treated as independent observations. The random intercept accounts for this by allowing each participant their own baseline level of performance, such that the comparison between agents is made relative to that individual rather than to the group mean. This isolates the effect of the agent's policy from the substantial variation in baseline competence between participants.

Given that the hypothesis requires the proposed agent to outperform each of the baselines, all three contrasts must reach significance for it to be supported. Requiring every test to succeed is a stricter criterion than requiring any one of them to succeed, so the risk of a false positive is already controlled and no correction for multiple comparisons is applied.

A secondary analysis will examine episodes separately by outcome, indicating whether an agent contributes primarily by maximising the score of successful attempts or by limiting the losses incurred during a crash. Finally, further analysis will investigate whether the agents' support varies across different user demographics. By comparing performance across age groups and levels of video game experience, this will reveal whether the cognitively constrained agent provides a universal benefit, or if it disproportionately aids less experienced players by compensating for a lack of pre-existing aptitude.

\section{Extended Results}
\label{app:extended}

Twenty-two metrics in total were analysed when comparing how similar the constrained and unconstrained agents were to human players, shown in Table \ref{tab:features}.

Three measures are reported for each, since no single one is adequate alone. The raw difference in group means is directly interpretable but scale-dependent, so cannot be compared across features. The Wasserstein distance normalised by group spread represents how far apart distributions are, but can be distorted by distributions with one long tail. The overlap measures the shared probability mass, but cannot distinguish magnitudes of separation if distributions share no overlap.

The mean press duration refers to the average amount of time an input is held down for on each instance of its use, with the main and side thrusters measured separately. The final displacement is the absolute horizontal distance from the landing pad at termination, with the tilt and speed measuring the angle and velocity at the last frame of the game. Idle refers to the proportion of frames throughout the episode where each or neither of the engines are inactive, with mean idle length referring to the average duration of intervals in which neither control is active. Tilt measures the root-mean-square tilt across the episode. Score refers to the total reward accumulated over the episode. Press duration CV is the coefficient of variation, computed by dividing the standard deviation of press lengths by their mean. Vertical speed is the mean descent rate across the episode. Inputs per second is the number of separate presses made each second, counted once per press regardless of how long it is held, and measured separately for the main and side thrusters. Episode length is the number of frames elapsed before the episode terminates. Action entropy refers to the Shannon entropy over the three lateral thruster states (off/left/right), measuring the shape of the whole action distribution. The reversal rate is the number of times left-to-right or right-to-left switches with a gap of 10 frames or less occur per second, and the burst rate measures occurrences of three or more direction changes within 25 frames, per second. The corrective latency counts the frames from the craft tilting past an angle of 0.1 radians off centre to when a thrust occurs to correct this, and the left-right balance measures the directional asymmetry between the left and right actions.

\begin{table}[H]
\centering
\caption{Individual feature comparison of agents with the human distribution. Overlap is the shared probability mass (higher is closer); distance is the normalised Wasserstein distance (lower is closer). Bold indicates the closer of the two agent types under each of the two distributional measures.}
\label{tab:features}
\small
\begin{tabular}{@{}lcccc@{}}
\toprule
& \multicolumn{2}{c}{Overlap} & \multicolumn{2}{c}{Wasserstein} \\
\cmidrule(lr){2-3} \cmidrule(lr){4-5}
Feature & Constrained & Unconstrained & Constrained & Unconstrained \\
\midrule
Mean press duration (side)        & \textbf{0.73} & 0.00 & \textbf{0.32} & 2.01 \\
Final speed                 & \textbf{0.77} & 0.50 & \textbf{0.72} & 1.07 \\
Final displacement          & \textbf{0.59} & 0.25 & \textbf{0.73} & 1.32 \\
Final tilt                  & \textbf{0.78} & 0.75 & \textbf{0.74} & 0.94 \\
Idle (side)      & 0.36 & \textbf{0.71} & 0.86 & \textbf{0.42} \\
Tilt (RMS)                  & \textbf{0.67} & 0.22 & \textbf{0.89} & 1.14 \\
Mean idle length              & \textbf{0.17} & 0.00 & \textbf{0.89} & 1.95 \\
Score                       & \textbf{0.43} & 0.05 & \textbf{0.90} & 1.39 \\
Idle (main)           & 0.29 & \textbf{0.56} & \textbf{0.91} & 1.02 \\
Press duration CV (side)     & \textbf{0.49} & 0.09 & \textbf{0.92} & 1.73 \\
Vertical speed              & \textbf{0.32} & 0.31 & \textbf{1.06} & 1.11 \\
Action entropy              & 0.26 & \textbf{0.42} & 1.21 & \textbf{0.50} \\
Idle (neither control)       & 0.24 & \textbf{0.61} & 1.22 & \textbf{0.35} \\
Press duration CV (main)     & \textbf{0.40} & 0.31 & 1.22 & \textbf{1.05} \\
Reversal rate               & 0.26 & \textbf{0.49} & 1.26 & \textbf{0.85} \\
Episode length              & \textbf{0.34} & 0.05 & \textbf{1.27} & 1.58 \\
Burst rate                  & \textbf{0.24} & 0.00 & \textbf{1.27} & 1.65 \\
Corrective latency          & 0.24 & \textbf{0.52} & 1.42 & \textbf{1.21} \\
Inputs per second (main)     & \textbf{0.19} & 0.00 & \textbf{1.44} & 1.69 \\
Inputs per second (side)     & \textbf{0.24} & 0.00 & \textbf{1.47} & 1.50 \\
Left/right balance          & 0.24 & \textbf{0.47} & 1.49 & \textbf{1.13} \\
Mean press duration (main)        & \textbf{0.19} & 0.02 & \textbf{1.50} & 1.95 \\
\bottomrule
\end{tabular}
\end{table}

\begin{table}[H]
\centering
\caption{Mean value and standard deviation of each feature by group, averaged within and then across controllers. Bold indicates the agent type nearer the human mean.
Reported alongside Table~\ref{tab:features}, since normalisation via pooled standard deviation can distort rankings for features with long tails.}
\label{tab:raw_features}
\small
\setlength{\tabcolsep}{6pt}
\renewcommand{\arraystretch}{1.15}
\begin{tabular}{@{}l c@{\hskip 10pt}c @{\hskip 20pt} c@{\hskip 10pt}c @{\hskip 20pt} c@{\hskip 10pt}c@{}}
\toprule
& \multicolumn{2}{c}{Human} & \multicolumn{2}{c}{Constrained} & \multicolumn{2}{c}{Unconstrained} \\
\cmidrule(lr){2-3} \cmidrule(lr){4-5} \cmidrule(lr){6-7}
Feature & \multicolumn{1}{c}{Mean} & \multicolumn{1}{c}{(SD)}
        & \multicolumn{1}{c}{Mean} & \multicolumn{1}{c}{(SD)}
        & \multicolumn{1}{c}{Mean} & \multicolumn{1}{c}{(SD)} \\
\midrule
Mean press duration, side (ms)   & 298.5 & (80.1)   & \textbf{279.8} & (98.2)  & 47.1 & (9.3) \\
Mean press duration, main (ms)   & 561.8 & (171.0)  & \textbf{353.2} & (49.3)  & 98.5 & (42.4) \\
Press duration CV, side          & 0.438 & (0.173)  & \textbf{0.533} & (0.088) & 0.982 & (0.196) \\
Press duration CV, main          & 0.537 & (0.084)  & \textbf{0.435} & (0.063) & 0.858 & (0.333) \\
Mean idle length (ms)            & 321.1 & (104.7)  & \textbf{196.1} & (192.5) & 49.2 & (13.0) \\
Inputs per second, side          & 0.940 & (0.245)  & \textbf{1.623} & (0.416) & 6.122 & (2.970) \\
Inputs per second, main          & 0.924 & (0.270)  & \textbf{1.422} & (0.284) & 4.766 & (1.687) \\
Burst rate (per s)               & 0.432 & (0.406)  & \textbf{1.391} & (0.690) & 9.733 & (4.344) \\
Reversal rate (per s)            & 0.144 & (0.125)  & \textbf{0.607} & (0.349) & 2.362 & (2.908) \\
Corrective latency (ms)          & 286.3 & (132.0)  & 130.5 & (46.5)  & \textbf{145.8} & (76.7) \\
Idle, side (fraction)            & 0.696 & (0.130)  & 0.565 & (0.167) & \textbf{0.714} & (0.108) \\
Idle, main (fraction)            & 0.569 & (0.049)  & \textbf{0.531} & (0.074) & 0.608 & (0.025) \\
Idle, neither control (fraction) & 0.452 & (0.077)  & 0.311 & (0.129) & \textbf{0.446} & (0.058) \\
Action entropy (bits)            & 0.997 & (0.128)  & 1.283 & (0.266) & \textbf{1.066} & (0.222) \\
Left/right balance               & 0.340 & (0.205)  & 0.101 & (0.036) & \textbf{0.170} & (0.075) \\
Tilt, RMS (rad)                  & 0.317 & (0.213)  & \textbf{0.199} & (0.044) & 0.159 & (0.018) \\
Vertical speed                   & 0.343 & (0.125)  & 0.217 & (0.112) & \textbf{0.342} & (0.026) \\
Final displacement               & 0.263 & (0.227)  & \textbf{0.202} & (0.073) & 0.055 & (0.022) \\
Final speed                      & 0.477 & (0.704)  & \textbf{0.189} & (0.223) & 0.001 & (0.007) \\
Final tilt (rad)                 & 0.367 & (0.630)  & \textbf{0.142} & (0.066) & 0.008 & (0.012) \\
Episode length (frames)          & 261.0 & (87.4)   & 397.8 & (89.0)  & \textbf{198.6} & (19.9) \\
Score                            & 77.3  & (211.0)  & \textbf{108.6} & (50.0)  & 286.3 & (8.2) \\
\bottomrule
\end{tabular}
\end{table}

\begin{figure}[H]
    \centering
    \includegraphics[width=\textwidth]{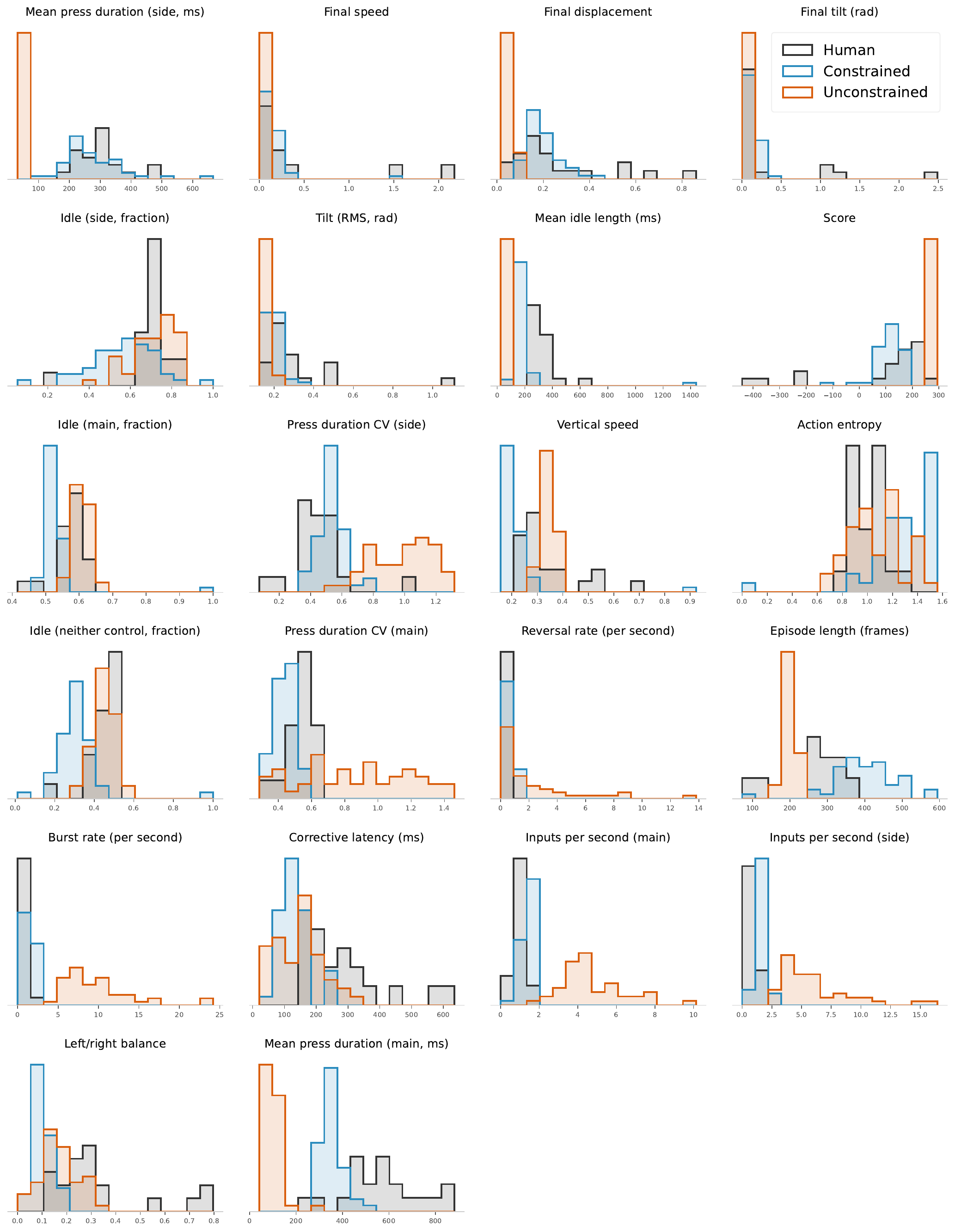}
    \caption{Distribution of behavioural features comparing human, constrained agent, and unconstrained agent groups.}
    \label{fig:behavioural_features}
\end{figure}

Confidence intervals throughout Section~\ref{sec:crossplay} are bootstrap percentiles over 5{,}000 resamples. The design has two sampling units rather than one: the ten independently trained replicates of each condition, and the fifteen held-out partners comprising each pool. Episodes are first collapsed to a mean for each (replicate, partner) pairing, since the episodes one policy plays against one partner are repeated measurements of that pairing rather than independent observations; resampling episodes directly would treat them as independent and yield intervals several times too narrow.

Because every replicate meets every partner, the two units are crossed rather than nested. Each bootstrap iteration therefore draws one set of replicates and one set of partners, with replacement, and applies the same partner draw across all replicates. Resampling partners independently within each replicate would average the partner variance component over the ten replicates and understate it by a factor of approximately $\sqrt{10}$.

Table~\ref{tab:ownpool} pairs by replicate such that the same replicates are drawn for the own-pool and held-out sides of each difference, so a draw containing unusually strong runs raises both sides together and run strength cancels. Table~\ref{tab:heldout} pairs by partner instead, drawing one partner set applied to both agents so that per-partner difficulty cancels, but resamples replicates independently for each agent, since the two methods' training runs have no correspondence on which to pair them.

All intervals are conditional on the fixed set of twenty evaluation seeds, which is held constant across every agent, partner and replicate.

\begin{table}[H]
\centering
\caption{Held-out advantage of the constrained agent over each baseline, restricted to the partner families neither agent trained on so that both are averaged over the same partners. Intervals are percentiles of a two-way cluster bootstrap over 5{,}000 resamples, paired by partner so that per-partner difficulty cancels from the difference, with replicates resampled independently for each agent.}
\label{tab:heldout}
\begin{tabular}{llccl}
\toprule
Comparison & Shared families & Advantage & 95\% CI & Significance \\
\midrule
vs PBT  & FCP, MARL & $-4.5$  & $[-11.4, +1.9]$ & Not significant \\
vs FCP  & PBT, MARL & $+4.0$  & $[-1.8, +10.0]$ & Not significant \\
vs MARL & PBT, FCP  & $+27.5$ & $[+8.7, +52.7]$ & Significant \\
\bottomrule
\end{tabular}
\end{table}

\section{Questionnaire Items}
\label{app:questionnaire}

Immediately after their final block of solo play, participants will complete a short questionnaire covering demographic and gaming-experience covariates followed by their subjective experience of each teaming agent, as described in Section~\ref{sec:study}. Agents are identified to participants only by a fixed colour label (White, Blue, Red, Yellow), held consistent across all participants. Items 6--9 refer back to the ranking given in Item 5, so that each participant's most- and least-successful agent is unambiguous both to them and in the resulting data.

\begin{enumerate}
    \item Participant number (free entry), used to link responses to the
    corresponding session log.
    \item Age, in one of seven bands: 18--25, 26--35, 36--45, 46--55,
    56--65, 66--75, 76+.
    \item Frequency of video game play, on a six-point scale from Daily
    to Never.
    \item Perceived difference between the four agents' cooperative
    ability, from 1 (no difference) to 10 (large difference).
    \item Ranking of the four agents by perceived success of the
    partnership (1 = most successful, 4 = least successful).
    \item Rated impact of the agent ranked first in Item 5 on the
    participant's score, from 1 (very detrimental) to 10 (very helpful).
    \item As Item 6, for the agent ranked last.
    \item Rated extent to which the agent ranked first in Item 5
    supported the participant's intended play style, from 1 (became a
    barrier to the participant's intentions) to 10 (understood and
    supported the participant's intentions).
    \item As Item 8, for the agent ranked last.
    \item Overall preference between solo play and play supported by a
    teaming agent (preferred alone / no preference / preferred with
    agent).
    \item Free-text comments on specific agents or the teaming
    experience in general.
\end{enumerate}

\section{Gameplay Information}
\label{app:instructions}

\begin{figure}[H]
    \centering
    \includegraphics[width=0.8\linewidth]{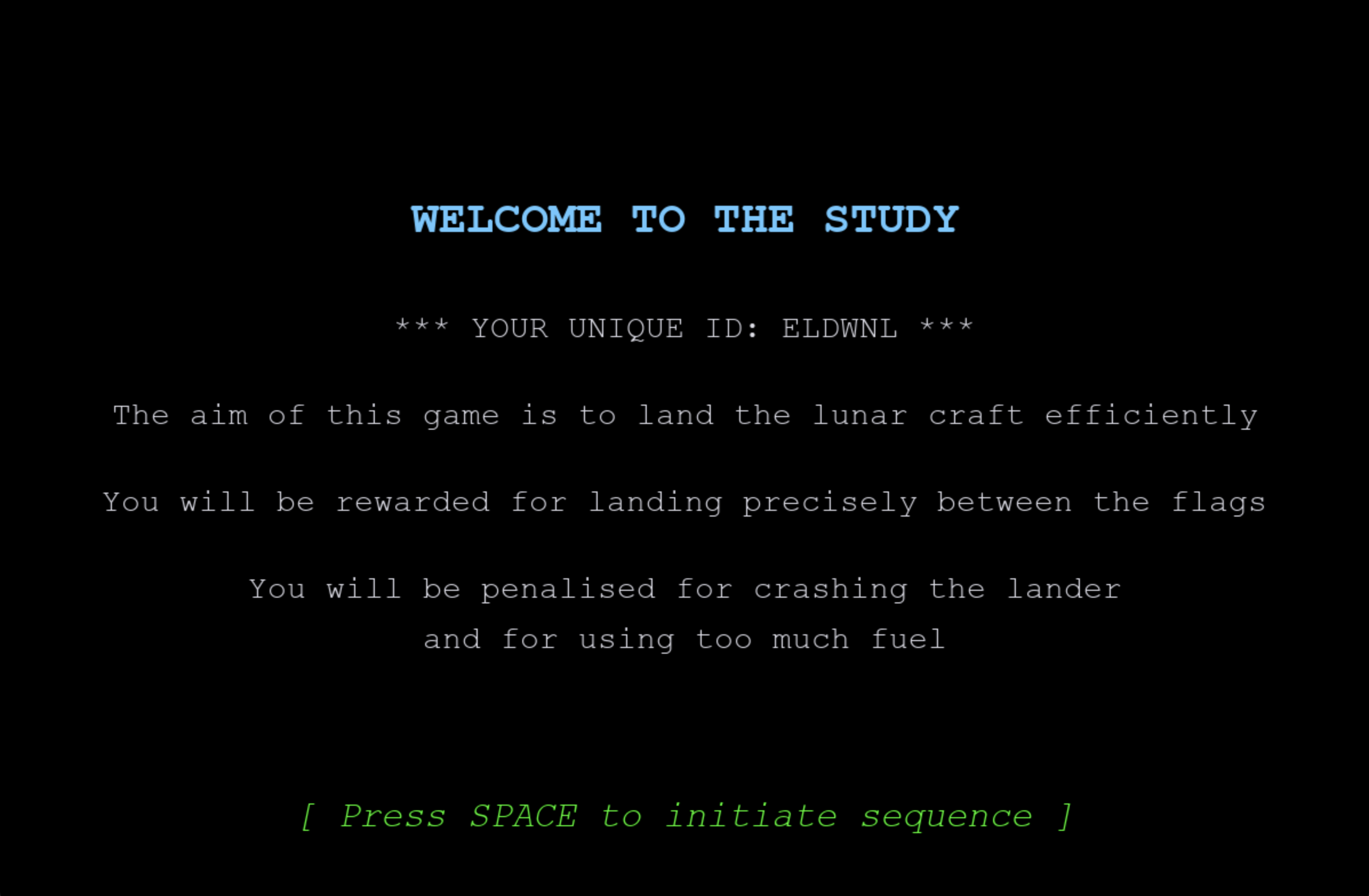}
    
    \vspace{1em}

    \includegraphics[width=0.8\linewidth]{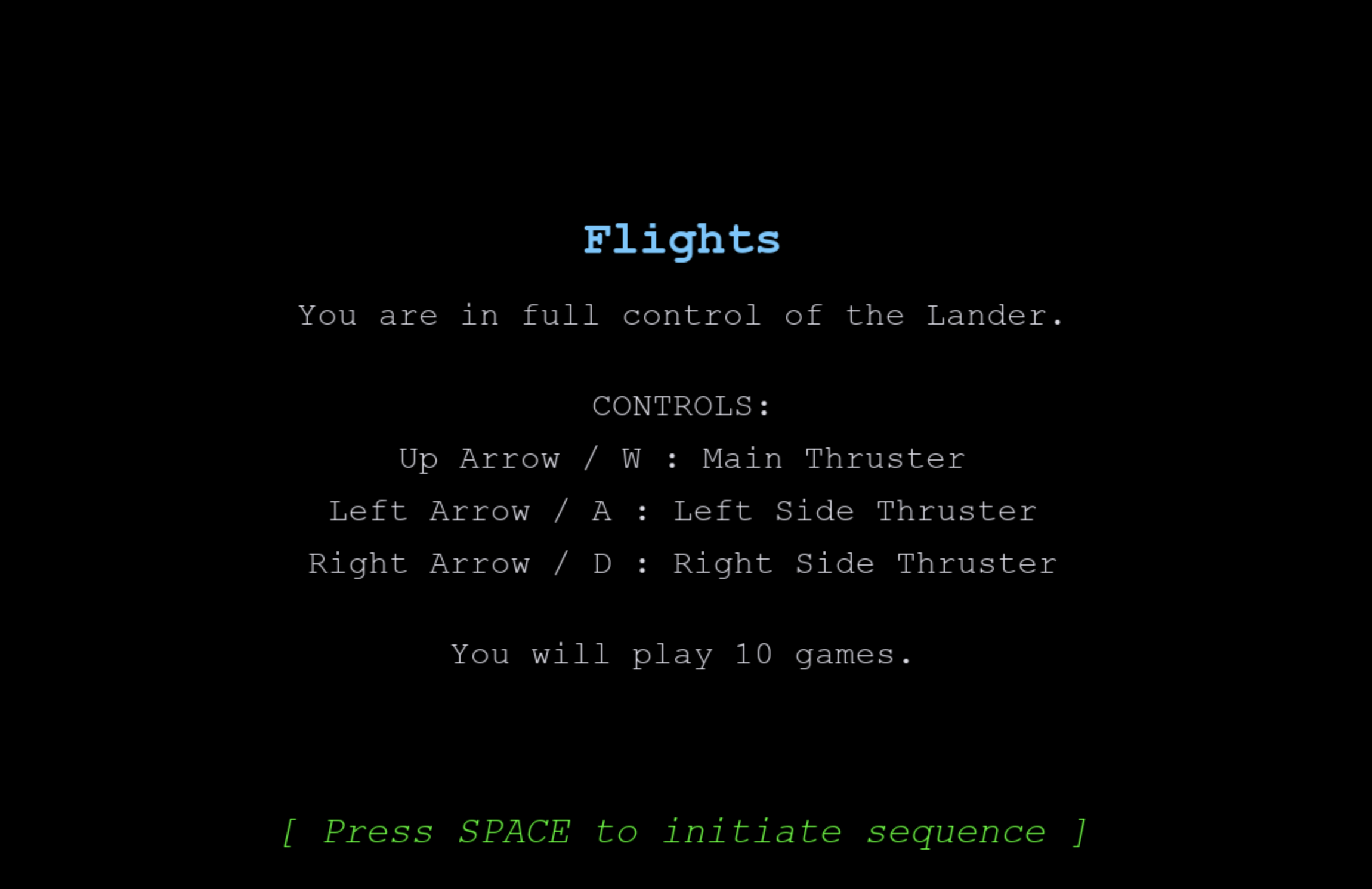}
    
    \caption{Instructions seen by participants at the start of the trial.}
    \label{fig:instructions}
\end{figure}

\end{document}